\documentclass[10pt]{article}
\usepackage[margin=1.5cm]{geometry}
\usepackage[title]{appendix}
\usepackage{tikz}
\usetikzlibrary{fadings}
\usetikzlibrary{shapes.multipart, shapes.geometric, arrows.meta}

\usetikzlibrary{shapes}
\usetikzlibrary{shapes.arrows}
\usetikzlibrary{shapes.misc}
\usetikzlibrary{shadows}
\usetikzlibrary{arrows}
\usetikzlibrary{positioning}
\usetikzlibrary{matrix}
\usetikzlibrary{calc}
\usepackage{pgfplots}
\usepackage{wrapfig}
\usepackage{subcaption}

\newtheorem{remark}{Remark}

\newtheorem{proposition}{Proposition}

\usepackage{amsmath, amssymb,mathtools}

\usepackage[utf8]{inputenc} % allow utf-8 input
\usepackage[T1]{fontenc}    % use 8-bit T1 fonts
\usepackage{hyperref}       % hyperlinks
\usepackage{url}            % simple URL typesetting
\usepackage{booktabs}       % professional-quality tables
\usepackage{amsfonts}       % blackboard math symbols
\usepackage{nicefrac}       % compact symbols for 1/2, etc.
\usepackage{microtype}      % microtypography
\usepackage{cleveref}       % smart cross-referencing
\usepackage{lipsum}         % Can be removed after putting your text content
\usepackage{graphicx}
\usepackage{doi}

\title{A Two-Step Rule for Backpropagation}

\author{{Ahmed Boughammoura}${}^*$}

\usepackage{stmaryrd}
\usepackage{nccmath}
\makeatletter
 \newcommand{\xmapsfrom}[2][]{%
 \ext@arrow3095\leftarrowfill@{#1}{#2}\mapsfromchar
}
\makeatother
\usetikzlibrary{decorations.pathreplacing}
    \usetikzlibrary{positioning}
     \usetikzlibrary{calc}
\usepackage{multicol}\usepackage{bm}\definecolor{ourspecialtextcolor}{rgb}{0.528, 0.471, 0.701} %
\usepackage{algorithm}
\usepackage[noend]{algpseudocode}
\DeclareCaptionSubType*{algorithm}

\algrenewcommand{\algorithmiccomment}[1]{\bgroup\hfill//~#1\egroup}
\algrenewcommand{\Return}{\State\textbf{return}\ }
\algnewcommand{\Save}{\State\textbf{save}\ }
\algnewcommand{\Load}{\State\textbf{load}\ }

\usepackage{xspace}

\begin{document}
\maketitle
%%%%%%%%%%%%%%%%%%%%%%%%%%%%%%%%%

%%%%%%%%%%%%%%%%%% Footnote%%%%%%%%%%%%%%%%%%%%%%%
\makeatletter{\renewcommand*{\@makefnmark}{}
\footnotetext{${}^*$ Department of Mathematics\\ \hspace*{0.75cm} Higher Institute of Informatics and Mathematics, Monastir, Tunisia\\ \hspace*{0.75cm} {Email: {ahmed.boughammoura@gmail.com}}\makeatother}
%%%%%%%%%%%%%%%%%%%%%%%%%%%%%%%%%%%%%%%%%%%%%%%%%

%%%%%%%%%%%%%%%%%%%%%%%%%%
\begin{abstract}
We present a simplified computational rule  for the  back-propagation formulas for artificial neural networks. In this work, we  provide  a  generic two-step rule    for the back-propagation algorithm in matrix notation. Moreover, this rule  incorporates both the forward and backward phases of the computations involved in the learning process.  Specifically, this recursive computing rule permits the propagation of the changes to all synaptic weights in the network, layer by layer, efficiently. In particular, we use this  rule to compute both the up  and down partial derivatives     of the cost function  of all the connections feeding into the output layer. 
\end{abstract}

% keywords can be removed
%\keywords{First keyword \and Second keyword \and More}
\section{Introduction}
~~~ An Artificial Neural Network (ANN) is a mathematical model which is intended to be a universal function approximator which learns from data (cf. McCulloch and Pitts, \cite{McCulloch1943}). 
In general, an ANN consists of a number of units called artificial neurons, which are  a composition of affine mappings, and non-linear (activation) mappings (applied element wise), connected by weighted connections and organized into layers, containing an input layer, one or more hidden layers, and an output layer.The neurons in an ANN can be connected in many different ways. In the simplest cases, the outputs from one layer are the inputs for the neurons in the next
layer. An ANN  is said to be a feedforward ANN, if outputs from one layer of neurons
are the only inputs to the neurons in the following layer. In a fully connected ANN, all neurons in one layer are connected to all neurons in the previous layer (cf. page 24 of \cite{Baldi2021}). An example of a fully connected feedforward network is presented in Figure \ref{fig:multilayer-perceptron}.
%%%%%%%%%%%%%%%%%%%%%%%%%%%%%%%%%%%%%%%%%%%%%%%%%%%%%%%%%%%%%%%%%%%%%%%%%%%%%%%%%%%%%%%%%%%%

In the present work we focus essentially on feed-forward artificial neural networks, with $L$ hidden layers  and a transfer (or activation) function $\sigma$, and the corresponding  supervised learning problem. Let us define a simple artificial neural network as follows:

\begin{equation}
\label{feedforward-formulas-scalar1}
X_{0}^{\mathrm{out}}= x,\ 
Y_{h}^{\mathrm{out}}=W_{h}.X_{h-1}^{\mathrm{out}},\  
  X_{h}^{\mathrm{out}}=\sigma(Y_{h}^{\mathrm{out}}), \ h=1,\cdots, L
\end{equation}
where $ x\in\mathbb{R}^n$ is the input to the network, $h$ indexes the hidden layer and $W_h$ is the weight matrix of the $h$-th hidden layer. In what follows we shall refer to the two equations of \eqref{feedforward-formulas-scalar1} as the two-step recursive forward formula. The two-step recursive forward formula is very useful in obtaining the outputs of the feed-forward deep neural networks.

A major empirical issue in the neural networks is to estimate the unknown parameters
$W_h$ with a sample of data values of targets and inputs. This estimation procedure is characterized by the recursive updating or the learning of estimated parameters. This algorithm is called the   backpropagation algorithm. 
 As reviewed by Schmidhuber \cite{Schmidhuber15}, back-propagation was introduced and developed  during the 1970’s and 1980’s and refined  by Rumelhart et al. \cite{rumelhart1986}). In addition, it is well known that  the most important algorithms of artificial neural networks training is the back-propagation algorithm. From mathematical point view,  back-propagation is a method  to minimize errors for a loss/cost function through gradient descent.  More precisely, an input data is fed to the network and forwarded through the so-called layers ; the produced output is then fed to the cost function to compute the    gradient of the associated error. The computed gradient is then back-propagated through the layers to update the weights by using the well known gradient descent algorithm.

As explained in \cite{rumelhart1986}), the goal of back-propagation is to compute the partial derivatives  of the cost function $J$. In this procedure,  each hidden layer $h$ is assigned an error term $\delta^h$. For each hidden layer, the error term $\delta^h$ is derived from error terms   $\delta^k,\ k=h+1,\cdots L$; thus the concept of error back-propagation. The output layer $L$ is the only layer
whose error term $\delta^L$ has no error dependencies, hence $\delta^L$  is then given by the following equation
\begin{equation}
\label{delta-L}
\delta^L=\frac{\partial J}{\partial W_L}\odot \sigma'(Y_{L}^{\mathrm{out}}),
\end{equation}
where $\odot$  denotes the element-wise matrix multiplication (the so-called Hadamard product, which is exactly the element-wise multiplication  $"*"$ in Python).
%%%%%%%%%%%%%%%%%%%%%%% 
For the error term $\delta^h$, this term is derived from matrix multiplying $\delta^{h+1}$  with the weight transpose matrix $(W_{h+1})\top$ and subsequently
multiplying (element-wise) the activation function derivative $\sigma'$ with respect to the preactivation $Y_{h}^{\mathrm{out}}$. Thus, one has the following equation
\begin{equation}
\label{delta-l}
\delta^h=(W_{h+1})^\top\delta^{h+1}\odot \sigma'(Y_{h}^{\mathrm{out}}),\ h=(L-1),\cdots, 1
\end{equation}
Once the layer error terms have been assigned, the partial derivative $\frac{\partial J}{\partial W_l}$ can be computed by
\begin{equation}
\label{delta-W}
\frac{\partial J}{\partial W_h}=\delta^{h+1}(X_{h}^{\mathrm{out}})^\top
\end{equation}
In particular, we deduce that the back-propagation algorithm is uniquely  responsible for computing weight partial
derivatives of $J$ by using the recursive equation \eqref{delta-l} with the initialization data given by \eqref{delta-L}.
%%%%%%%%%%%%%%%%%%%%%%%%%%%%%%%%%%%%%%%%%%%%%%%%%%%%%%%%%%%%%%%%%%%%%%%%%%%%%%%%%%%%%%%%%%%%
The key question to which we address ourselves in the present work is the following: how could one reformulate the back-propagation in a similar manner as in the forward pass ? Equivalently, how could one reformulate the back-propagation  in two-step recursive backward formula as in \eqref{feedforward-formulas-scalar1} ?
 
 In order to provide a first answer to this question, we shall introduce the following up and down delta's terms
 \begin{equation}
\label{delta-X} \delta_L^{\mathrm{up}}:=\frac{\partial J}{\partial X_{L}^{\mathrm{out}}},\ 
\delta_h^{\mathrm{down}}=\delta_h^{\mathrm{up}}\odot \sigma'( Y_{h}^{\mathrm{out}}) ,\ \delta_{h-1}^{\mathrm{up}}= (W_h)^\top\delta_h^{\mathrm{down}},\ h=L,\cdots, 1
\end{equation}
Once the  $\delta_h^{\mathrm{down}}$ term  have been computed, the partial derivative $\frac{\partial J}{\partial W_h}$ can be evaluated by
\begin{equation}
\label{delta-W:2}
\frac{\partial J}{\partial W_h}= \delta_h^{\mathrm{down}}(X_{h-1}^{\mathrm{out}})^\top
\end{equation}
Now, we shall give an answer by proving in the section 3 that one has the   two-step recursive backward formula given by \eqref{delta-X}. To end this Introduction, let us mention that an earlier version of this manuscript was submitted as a preprint \cite{bougham2023}.
%%%%%%%%%%%%%%%%%%%%%%%%%%%%%%%%%%%%%%%%%%%%%%%%%%%%%%%%%%%%%%%%%%%%%%%

The rest of the paper is organized as follows.  
%Section 2 gives an overview of the Event-B language. 
Section 2 outlines some notations, setting and ANN framework. Section 3 state and proof the main mathematical result of this work.
Section 4 application of this method to study some simple cases. In Section 5 conclusion, related works and mention
future work directions.

%%%%%%%%%%%%%%%%%%%%%%
\section{Notations, Setting  and the ANN} 
%%%%%%%%%%%%%%%%%%%%%%%%%%%%%%%%%%%%%%%%%%%%%%%%%%%%%%%%%%%%
~~~Let us now precise some notations. Firstly, we shall denote  any vector $X \in \mathbb{R}^n$, is considered as columns $X=\left(X_1,\cdots, X_{n} \right)^\top$ and for any  family of transfer functions $\sigma_i: \mathbb{R}\rightarrow \mathbb{R},\ i=1,\cdots, n$, we shall introduce the coordinate-wise map $\sigma : \mathbb{R}^n \rightarrow \mathbb{R}^n$   by the following formula
\begin{equation}
\label{transfer-formulas}
\sigma(X):=\left(\sigma_1(X_1),\cdots, \sigma_n(X_{n}) \right)^\top.
\end{equation}
This map   can be considered as an
“operator” Hadamard multiplication of columns $\sigma=\left(\sigma_1,\cdots, \sigma_{n} \right)^\top$ and 
$X=\left(X_1,\cdots, X_{n} \right)^\top$, i.e., 
$\sigma(X)=\sigma\odot X.$
%
%

 %=======================================
Secondly, we shall need to recall some useful multi-variable
functions derivatives notations. For any $n,m\in \mathbb{N}^*$ and any differentiable function  with respect to the variable $x$
\begin{equation}
F:\ \mathbb{R}\ni x\mapsto F(x)=\Bigl(F_{ij}(x)\Bigr)_{\substack{1\leq i\leq m \\ 1\leq j\leq n }}  \in \mathbb{R}^{m\times n}
\end{equation} we use the following notations associated to the partial derivatives of  $F$ with respect to $x$ 
\begin{equation}
	 \frac{\partial F}{\partial x} =
		\left(
		\frac{\partial F_{ij}(x)}{\partial x} 
	\right)_{\substack{1\leq i\leq m \\ 1\leq j\leq n }} 
	\end{equation} 
	
In adfdition, for any $n,m\in \mathbb{N}^*$ and any differentiable function  with respect to the matrix variable
\begin{equation}
F:\ \mathbb{R}^{m\times n}\ni X=\Bigl(X_{ij}\Bigr)_{\substack{1\leq i\leq m \\ 1\leq j\leq n }}\mapsto F(X) \in \mathbb{R}
\end{equation}

 we shall use the so-called  \textbf{denominator layout} notation (see page 15 of \cite{Ye2022}) for the partial derivative of  $F$ with respect to the matrix  $X$ 
\begin{equation}
	 \frac{\partial F}{\partial X} =
		\left(
		\frac{\partial F(X)}{\partial X_{ij}} 
	\right)_{\substack{1\leq i\leq m \\ 1\leq j\leq n }} 
	\end{equation}

In particular, this notation leads to the following useful formulas: for any $q\in\mathbb{N}^*$ and any matrix $W\in \mathbb{R}^{q\times m}$ we have
 \begin{equation}\label{usefl-x}
	 \frac{\partial (WX)}{\partial X} =W^\top, 
	\end{equation}
	when $X\in \mathbb{R}^{ n}$ with $X_n=1$ one has
 \begin{equation}\label{usefl-w-sharpe}
	 \frac{\partial (WX)}{\partial X} =W_\sharp^\top
	\end{equation} 
	where $W_\sharp$ is the matrix $W$ whose last column is removed (this formula  is highly useful in practice). 
Moreover,	for any   matrix $X\in \mathbb{R}^{m\times n}$ we have
 \begin{equation}\label{usefl-w}
	 \frac{\partial (WX)}{\partial W} =X^\top.
	\end{equation}
	
Then, by the chain rule one has
for any $q,n,m\in \mathbb{N}^*$ and any differentiable function  with respect to the matrices variables $W,X$ :
\begin{equation*}
F:\ (W, X)\mapsto Z:=WX\in \mathbb{R}^{q\times n}\mapsto F(Z) \in \mathbb{R}
\end{equation*}
	\begin{equation}\label{chain:}
	 \frac{\partial F}{\partial X} =
				W^\top\frac{\partial F}{\partial Z} \ 
				\mathrm{and}\ 
				\frac{\partial F}{\partial W} =
				\frac{\partial F}{\partial Z}X^\top.
	\end{equation}
Furthermore, for any   differentiable function  with respect to  $Y$
\begin{equation*}
F:\ \mathbb{R}^n\ni Y\mapsto X:=\sigma(Y)\in \mathbb{R}^{n}\mapsto F(X) \in \mathbb{R}
\end{equation*} we have
	\begin{equation}\label{chain:2}
	 \frac{\partial F}{\partial Y} =
				 \frac{\partial F}{\partial X}\odot\sigma'(Y).
	\end{equation}	
%%%%%%%%%%%%%%%%%%%%%%%%%%%%%%%%%%%%%%%%%%%%%%%%%%%%%%%%%%%%%%%%%%%%
Throughout this paper, we consider layered feedforward neural networks and supervised learning tasks. Following \cite{Baldi2021} (see (2.18) in page 24), we will denote such an architecture by
%%%%%%%%%%%%%%%%%%%%%%%%%%%%%%%%%%%%%%%%%%%%%%%%%%%%%%%%%%%%%%%%
\begin{equation}
\label{architectue}
A[N_0, \cdots, N_h,\cdots, N_L]
\end{equation}
where $N_0$ is the size of the input layer, $N_h$ is the size of hidden layer $h$,
and $N_L$ is the size of the output layer; $L$ is defined as the depth of the ANN , then the neural network is called as Deep Neural Network (DNN).
 We assume that the layers are fully
connected, i.e.,  neurons between two adjacent layers are fully pairwise connected, but neurons within a single layer share no connections. 
%
%%%%%%%%%%%%%%%%%%%%%%%%%%%%%%%%%%%%%%%%%%%%%%%%%%%%%%%%%%%%%

%
%
\usetikzlibrary{fadings}
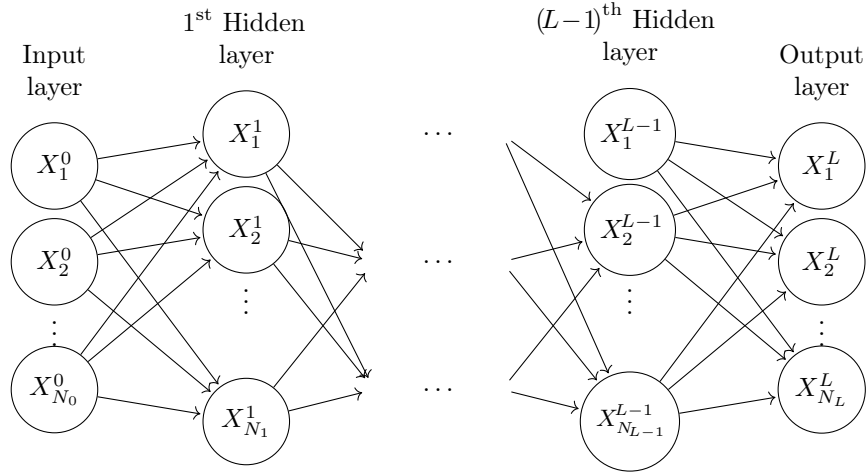
\begin{figure}[htp]
	\centering
	\begin{tikzpicture}[shorten >=1pt,scale=0.85]]
		\tikzstyle{unit}=[draw,shape=circle,minimum size=1.15cm]
		%\tikzstyle{hidden}=[draw,shape=circle,fill=black!25,minimum size=1.15cm]
		\tikzstyle{hidden}=[draw,shape=circle,minimum size=1.15cm]
		\node[unit](x0) at (0,3.5){$X_1^0$};
		\node[unit](x1) at (0,2){$X_2^0$};
		\node at (0,1){\vdots};
		\node[unit](xd) at (0,0){$X_{N_0}^0$};
		\node[hidden](h10) at (3,4){$X_1^{1}$};
		\node[hidden](h11) at (3,2.5){$X_2^{1}$};
		\node at (3,1.5){\vdots};
		\node[hidden](h1m) at (3,-0.5){$X_{N_{1}}^{1}$};
		\node(h22) at (5,0){};
		\node(h21) at (5,2){};
		\node(h20) at (5,4){};
		\node(d3) at (6,0){$\ldots$};
		\node(d2) at (6,2){$\ldots$};
		\node(d1) at (6,4){$\ldots$};
		\node(hL12) at (7,0){};
		\node(hL11) at (7,2){};
		\node(hL10) at (7,4){};
		\node[hidden](hL0) at (9,4){$X_1^{\!L-1\!}$};
		\node[hidden](hL1) at (9,2.5){$X_2^{\!L-1\!}$};
		\node at (9,1.5){\vdots};
		\node[hidden](hLm) at (9,-0.5){\small{$X_{N_{\!L-1\!}}^{\!L-1\!}$}};
		\node[unit](y1) at (12,3.5){$X_1^{L}$};
		\node[unit](y2) at (12,2){$X_2^{L}$};
		\node at (12,1){\vdots};	
		\node[unit](yc) at (12,0){$X_{N_L}^{L}$};
		\draw[->] (x0) -- (h10);
\draw[->] (x0) -- (h11);
		\draw[->] (x0) -- (h1m);
\draw[->] (x1) -- (h10);
		\draw[->] (x1) -- (h11);
		\draw[->] (x1) -- (h1m);
		\draw[->] (xd) -- (h11);
\draw[->] (xd) -- (h10);
		\draw[->] (xd) -- (h1m);
		\draw[->] (hL0) -- (y1);
		\draw[->] (hL0) -- (yc);
		\draw[->] (hL0) -- (y2);
		\draw[->] (hL1) -- (y1);
		\draw[->] (hL1) -- (yc);
		\draw[->] (hL1) -- (y2);
		\draw[->] (hLm) -- (y1);
		\draw[->] (hLm) -- (y2);
		\draw[->] (hLm) -- (yc);
		\draw[->,path fading=east] (h10) -- (h21);
		\draw[->,path fading=east] (h10) -- (h22);	
		\draw[->,path fading=east] (h11) -- (h21);
		\draw[->,path fading=east] (h11) -- (h22);
		\draw[->,path fading=east] (h1m) -- (h21);
		\draw[->,path fading=east] (h1m) -- (h22);
		\draw[->,path fading=west] (hL10) -- (hL1);
		\draw[->,path fading=west] (hL11) -- (hL1);
		\draw[->,path fading=west] (hL12) -- (hL1);
		\draw[->,path fading=west] (hL10) -- (hLm);
		\draw[->,path fading=west] (hL11) -- (hLm);
		\draw[->,path fading=west] (hL12) -- (hLm);
\node[above=5pt of x0, align=center](x0) {Input\\ layer};
		\node[above=5pt of h10, align=center] (h10) {${\!1}^{\text{st}}$ Hidden \\ layer};
		\node[above=5pt of hL0, align=center] (hL0){${\!\!(\!L\!-\!1\!)}^{\text{th}}$ Hidden \\ layer};
		\node[above=5pt of y1, align=center] (y1){Output\\ layer};
	\end{tikzpicture}
	\caption[Example of a $(L+1)$-layer perceptron.]{Example of an $A[N_0,\cdots, N_L]$ architechture.}
	\label{fig:multilayer-perceptron}
\end{figure}

For the rest of the paper, we will adopt some simplified notations by replacing some subscripts and superscripts.

~~~Now, let $\alpha_{ij}^h$ denote the weight connecting neuron $j$ in layer $h-1$ to neuron $i$ in hidden layer $h$ and let the associated transfer function denoted $\sigma_i^h$. 
%Without loss of generality, we shall assume that  for any $i$ and $h$,  $\sigma_i^h=\sigma$.
In general, in the application two different passes of computation are distinguished. The first pass is referred to as the forward pass, and the second is referred to as the backward pass. In the forward pass, the synaptic weights remain fixed throughout the network, and  the output  $X_i^h$ of neuron $i$ in hidden layer $h$ is computed by the following recursive-coordinate form :
\begin{equation*}
\label{feedforward-formulas-scalar}
X_i^h:=\sigma_i^h(Y_i^{h})\quad\mathrm{where}\quad Y_i^{h}:=\sum_{j=1}^{N_{h-1}}\alpha_{ij}^h X_j^{h-1}
\end{equation*}

In two-step recursive-matrix form, one may rewrite the above formulas as
\begin{equation*}
\label{feedforward-formulas-matrix}
X^h=\sigma^h(Y^{h})\  \mathrm{where}\  Y^{h}=W^hX^{h-1},
 \sigma^h:=(\sigma_1^h,\cdots,\sigma_{N_h}^h)^\top,\  W^h:=(\alpha_{ij}^h)\in 
\mathbb{R}^{{N_h}\times{N_{h-1}}}.
\end{equation*}
\begin{remark}\label{rem1}
{}\noindent

It is crucial to remark that, if we impose the following setting on $X^h, W^h$ and $\sigma_{N_h}^h$:
\begin{enumerate}
\item  all input vectors have the form $X^h = [X^h_1 , \cdots , X^h_{{N_h}-1}, 1]^\top$ for all $0\leq h\leq (L-1)$;
%\item  the last rows of all matrices $W^h$ for all $1\leq h \leq (L-1)$ have the form $[0, \cdots, 0, 1]$ ;
\item the last functions $\sigma_{N_h}^h$  in the columns $\sigma^h$ for all $1\leq h \leq (L-1)$ are   constant functions equal to $1$.
\end{enumerate}
Then, the $A[N_0,\cdots, N_L]$ neural network will be equivalent to a $(L-1)$-layered affine
neural network with $(N_0-1)$-dimensional input and $N_L$-dimensional output. Each hidden layer $h$ will contain $(N_h-1)$ “genuine” neurons and one (last) “formal”, associated to the bias; the last column of the matrix $W^h$ will be the bias vector for the $h$-th layer (For more details, see the examples given in Section 4). 
\end{remark}

%%%%%%%%%%%%%%%%%%%%%%%%%%%%%%%%%%%%%%%%%%%%%%%%%%%%%%%%%%%%%%
This forward pass computation between the two adjacent layers $h-1$ and $h$ may be represented mathematically as the composition of the following two maps:

\begin{align*}
{}\hskip1em  \mathbb{R}^{N_{h-1}} 
&\xrightarrow[\hskip2em]{}
\mathbb{R}^{N_{h}}\ \xrightarrow[\hskip2em]{}
\mathbb{R}^{N_{h}}
 \\
X^{h-1}
&\xmapsto[]{W^h}  
{W^h}X^{h-1}=Y^h\ \xmapsto[]{\sigma^h}  
\sigma^h(Y^h)=X^h
\end{align*}
%%%%%%%%%%%%%%%%%%%%%%%%%%%%%%%%%%%%%%%%%%%%%%%%%%%%%%%%%%%%%%%%%%
%%%%%%%%%%%%%%%%%%%%%%%%%%%%%%%%%%%%%%%%%%%
It could be presented as a simple  mapping diagram with $X^{h-1}$ as input and  the corresponding successive preactivation and activation  $Y^h={W^h}X^{h-1},X^h=\sigma^h(Y^h)$ as outputs (see Figure \ref{fig:forward-pass}).
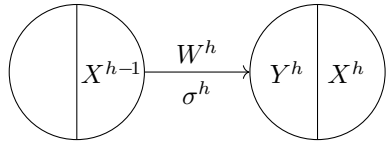
\begin{figure}[htp]
	\centering
\begin{tikzpicture}[xscale=2,yscale=1.0]
    \tikzstyle{neuron}=[circle,draw=black,minimum size=30pt,inner sep=0pt]
    \tikzstyle{neuron2}=[%
                       circle split,
                       %rectangle split horizontal,
                       %rectangle split parts=2,
                      % rounded corners=18pt,
                      % minimum size=37pt,
                      % minimum width=1.5cm,
                       text width=1.5cm,
                       draw=black,
                       rotate=90
                       ]
    \node[neuron2] (I1) at (0,5) {
       \nodepart[text width=1cm]{one}
       \nodepart[text width=1cm]{two} } ;     
     \node at (I1) {${}$\hspace{6.5ex}$\!X^{h-\!1}$} ;   
     \node[neuron2] (H1) at (1.6,5) {
       %\nodepart[text width=1cm]{one}
      % \nodepart[text width=1cm]{two} 
       } ;     
     \node at (H1) {$Y^h$\hspace{2ex}$X^h$} ;    
    \draw[->] (I1) -- (H1) node[midway,above] {$W^h$}  node[midway,below] {$\sigma^h$} ;                   
%
%\draw[->] (H1) --(3,5);                   
\end{tikzpicture}
\caption[ ]{Mapping diagram associated to the forward pass between two adjacent layers.}
	\label{fig:forward-pass}
\end{figure}

%%%%%%%%%%%%%%%%%%%%%%%%%%%%%%%%%%%%%%%%%%%%%%%%%%%%%%%%%%%%%
As consequence, the simplest neural network can be defined as a sequence of matrix multiplications and non-linearities:
$$X^0  = x,\ Y^h= W^hX^{h-1},\ X^h  = \sigma^h(Y^h),\ h  =  1,2,\cdots, L.
 $$
where $x\in \mathbb{R}^{N_0}$ is the input to the network, $h$ indexes the layer and $W^h$ is the weight matrix
of the $h$-th layer. To optimize the neural network, we compute the partial derivatives of the cost $J(.)$ w.r.t. the weight matrices $\frac{\partial J(.)}{\partial W^h}$. This quantity can be computed by making use of the chain rule in the back-propagation algorithm. To compute the partial derivative with respect to the matrices variables $\{{X^h},{Y^h}, {W^h}\}$, we put 
\begin{equation}
\label{notations}
\delta_h^{\mathrm{up}} =\frac{\partial J(.)}{\partial {X^h}},\ 
\delta_h^{\mathrm{down}}  =\frac{\partial J(.)}{\partial {Y^h}},\
\delta_{W^{h}}  =\frac{\partial J(.)}{\partial {W^h}}.
\end{equation}

Now, by using the two-step rule for back-propagation, introduced in the  previous section, one could rewrite the backward propagated values of the partial  derivatives of $J$ w.r.t. weight as follows :

\begin{equation}
\label{tsr1}
\delta_L^{\mathrm{up}}   =  \displaystyle\frac{\partial J(.)}{\partial X^L},\ 
\delta_h^{\mathrm{down}}  =  \displaystyle\delta_{h}^{\mathrm{up}}\odot\sigma'(Y^h),\ 
\delta_{W^{h}}  =  \displaystyle \delta_{h}^{\mathrm{down}}(X^{h-1})^\top
,\ 
\delta_{h-1}^{\mathrm{up}} =  \displaystyle({W^{h}})^\top \delta_{h}^{\mathrm{down}},\ 
h  =    L,\cdots, 2.
 \end{equation}

%%%%%%%%%%%%%%%%%%%%%%%%%%%%%%%%%%%%%%%%%%%%%%%%%%%%%%%%%

The backward   computation between the two adjacent layers $h$ and $h-1$ may be represented mathematically as follows:
%%%%%%%%%%%%%%%%%%%%%%%%%%%%%%%%%%%%%%%%%%%%%%%%%%%%%%
%%%%%%%%%%%%%%%%%%%%%%%%%%%%%%%%%%%%%%%%%%%%%%%%%%%%%%%%%%%%%%%%

\begin{align*}
\mathbb{R}^{N_{h-1}}\times (
\mathbb{R}^{N_{h}}\times \mathbb{R}^{N_{h-1}})
&\xleftarrow[\hskip7.5em]{}
 %
%\begin{bmatrix}
\mathbb{R}^{N_{h}}
%\end{bmatrix}
\xleftarrow[\hskip7.5em]{}
\mathbb{R}^{N_{h}}\qquad
 \\
\Bigl(\delta^{\mathrm{up}}_{h-1},
{\delta_{W^h}}
\Bigr)=\Biggl(\underbrace{(W^h)^\top}_{N_{h-1}\times N_h}\underbrace{\delta^{\mathrm{down}}_{h}}_{N_{h}\times 1}, 
\underbrace{\delta^{\mathrm{down}}_{h}}_{N_{h}\times 1}\underbrace{(X^{h-1})^\top}_{1\times N_{h-1}}
\Biggr) 
&\xmapsfrom[(W^{h})^\top{(.)}]{(.)(X^{h-1})^\top}   
{\delta^{\mathrm{down}}_{h}}=\delta^{\mathrm{up}}_{h}\odot{\sigma^h}'(Y^h)
\xmapsfrom[{}]{(\odot){\sigma^h}'(Y^h)}   
{\delta^{\mathrm{up}}_{h}}
\end{align*}
%%%%%%%%%%%%%%%%%%%%%%%%%%%%%%%%%%%%%%%%%%%

The simple  mapping diagram below shows the two-step rule for computing the partial derivatives of the cost function w.r.t. weights (see Figure \ref{fig:backward-pass}).
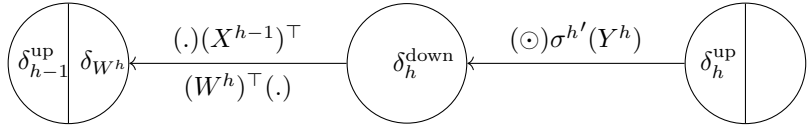
\begin{figure}[htp]
	\centering
\begin{tikzpicture}[xscale=3,yscale=1.0]
    \tikzstyle{neuron}=[circle,draw=black,minimum size=30pt,inner sep=16pt]
    \tikzstyle{neuron2}=[%
                       circle split,
                       %circle split horizontal,
                       %circle split parts=2,
                       %rounded corners=18pt,
                       minimum size=1pt,
                       minimum width=0.1cm,
                       text width=1.3cm,
                       draw=black,
                       rotate=90
                       ]
    \node[neuron2] (I1) at (0,5) {
       \nodepart[text width=0.2cm]{one}
       \nodepart[text width=1.5cm]{two} 
       %\nodepart[text width=1cm]{three}
       } ;     
     \node at (I1) {$\ \delta^{\mathrm{up}}_{h-1}$\hspace{0.1ex} $\delta_{W^{h}}$} ;   
 \node[neuron] (I2) at (1.5,5) {
       %\nodepart[text width=0.2cm]{one}
      % \nodepart[text width=1.5cm]{two} 
       %\nodepart[text width=1cm]{three}
       } ;     
     \node at (I2) {{}\hspace{3ex} $\delta^{\mathrm{down}}_{h}\ $} ; 
     \node[neuron2] (H1) at (3,5) {
       \nodepart[text width=0.2cm]{one}
       \nodepart[text width=1.5cm]{two} 
       %\nodepart[text width=1cm]{three} 
       } ;     
     \node at (H1) {$\quad\delta^{\mathrm{up}}_{h}$\hspace{7ex} } ;    
 \draw[<-] (I2) -- (H1) node[midway,above] {${(\odot)\sigma^h}'(Y^h)$}  node[midway,below] {} ;  
    \draw[<-] (I1) -- (I2) node[midway,above] {${(.)(X^{h-1})^\top}$}  node[midway,below] {$(W^h)^\top{(.)}$} ;                   
%
%\draw[->] (H1) --(3,5);                   
\end{tikzpicture}
\caption[ ]{Mapping diagram associated to the two-step backward pass between two adjacent layers.}
	\label{fig:backward-pass}
\end{figure}

%%%%%%%%%%%%%%%%%%%%%%%%%%%%%%%%%%%%%%%%%%%%%%%%%%%%%%%%%%%%%%%%%%%%%%%%%%%%%%%%%%%
%

One may  refine the above diagram  to show the similarity between both the forward and backward two-step passes as follows 
\begin{figure}[htp]
	\centering
\begin{tikzpicture}[xscale=2.5,yscale=1.0]
    \tikzstyle{neuron}=[circle,draw=black,minimum size=30pt,inner sep=0pt]
    \tikzstyle{neuron2}=[%
                       circle split,
                       %rectangle split horizontal,
                       %rectangle split parts=2,
                      % rounded corners=18pt,
                      % minimum size=37pt,
                      % minimum width=1.5cm,
                       text width=1.5cm,
                       draw=black,
                       rotate=90
                       ]
    \node[neuron2] (I1) at (0,5) {
       \nodepart[text width=1cm]{one}
       \nodepart[text width=1cm]{two} } ;     
     \node at (I1) {$\delta^{\mathrm{up}}_{h-\!1}$\hspace{1.ex}$\ \delta^{\mathrm{down}}_{h}$} ;   
     \node[neuron2] (H1) at (1.6,5) {
       %\nodepart[text width=1cm]{one}
      % \nodepart[text width=1cm]{two} 
       } ;     
     \node at (H1) {${}$\hspace{2ex}$\delta^{\mathrm{up}}_{h}\quad\qquad$} ;    
    \draw[<-] (I1) -- (H1) node[midway,above] {${(\odot)\sigma^h}'(Y^h)$}  node[midway,below] {$(W^h)^\top{(.)}$} ;                   
%
%\draw[->] (H1) --(3,5);                   
\end{tikzpicture}
\caption[ ]{Refined mapping diagram associated to the two-step backward pass between two adjacent layers.}
	\label{fig:backward-pass2}
\end{figure}
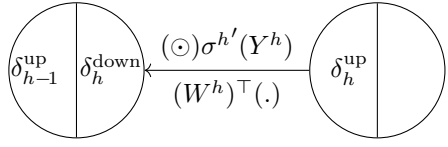

Note that Figure \ref{fig:forward-pass}  and Figure \ref{fig:backward-pass2} are adjoint to each other in both  computational phases. Moreover, one could combine the forward and backward passes by the following diagram, which shows clearly the two-step rule for the entire  back-propagation process (see Figure \ref{fig:forward-backward-pass}).
%%%%%%%%%%%%%%%%%%%%%%%%%%%%%%%%%%%%%%%%%are adjoint to each other%%%%%%%%%%%%%%%%%%%%%%%%%%%%%%%%%%%%%%%%%
\medskip

 \vspace*{2cm}
\begin{figure}[htp] 
    	\centering
%%%%%%%%%%%%%%%%%%%%%%%%%%%%%%%%%%%%%%%%%%%%%%%%%%%%%%%%%%%%%%%%%%%%%%
\begin{tikzpicture}[thick,   transform canvas={scale=0.86}]

%Nodes
\matrix[row sep=2cm,column sep=2.cm] (m1) {
    \node [align=center, xshift=1.5cm] (node21) {$Y^{1}$}; &
    \node [align=center, xshift=1.5cm] (node22) {$Y^{2}$}; &
    \node [align=center, xshift=1.cm] (node23) {$Y^{L-1}$}; 
&
    \node [align=center, xshift=3.5cm] (node24) {$Y^{L}$}; %&
    \\
    \node [align=center, xshift=.1cm] (node31) {$\delta^{\mathrm{up}}_{1}$}; &
    \node [align=center, xshift=.1cm] (node32) {$\delta^{\mathrm{up}}_{2}$}; &
    \node [align=center, xshift=.1cm] (node33) {$\delta^{\mathrm{up}}_{L-1}$}; &
    \node [align=center, xshift=5.5cm] (node34) {$\delta^{\mathrm{up}}_{L}$}; 
    \\
   };
%%%%%%%%%%%%%%%%%%%%%%%%%%%%%%%%%%%%%%%%%%%%%%%%%%%%%%%%%
\node [left of = node31, node distance = 1.9cm] (node20_31) {$\delta^{\mathrm{down}}_{1}$};
\node [left of = node32, node distance = 2.5cm] (node20_32) {$\delta^{\mathrm{down}}_{2}$};
\node [left of = node33, node distance = 2.5cm] (node20_33) {$\delta^{\mathrm{down}}_{L-1}$};
\node [left of = node34, node distance = 2.5cm] (node20_34) {$ \delta^{\mathrm{down}}_{L}$};
%%%%%%%%%%%%%%%%%%%%%%%%%%%%%%%%%%%%%%%%%%%%%%%%%%%%%%%%%
\node [below =1cm of node20_31, node distance = 2.9cm] (node10_31) {$\delta_{W^{1}}$};
\node [below =1cm of node20_32, node distance = 2.5cm] (node10_32) {$\delta_{W^{2}}$};
\node [below =1cm of node20_33, node distance = 2.5cm] (node10_33) {$\delta_{W^{L-1}}$};
\node [below =1cm of node20_34, node distance = 2.5cm] (node10_34) {$ \delta_{W^{L}}$};
%%%%
\draw [-Triangle] (node20_31) -- (node10_31);
\draw [-Triangle] (node20_32) -- (node10_32);
\draw [-Triangle] (node20_33) -- (node10_33);
\draw [-Triangle] (node20_34) -- (node10_34);
%%%%%%%%%%%%%%%%%%%%%%%%%%%%%%%%%%%%%%%%%%%%%%%%%%%%%%%%
\node [left of = node21, node distance = 3.3cm] (node20) {$ X^{0}=x$};
\node [right of = node21, node distance = 2cm] (node21_1) {$X^{1}$};
\node [right of = node22, node distance = 1.6cm] (node22_1) {$X^{2}$};
\node [right of = node23, node distance = 1.5cm] (node23_1) {$X^{L-1}$};
\node [right of = node24, node distance = 1.8cm] (node24_1) {$X^{L}$};

\draw [-Triangle] ([xshift=1.5cm]node20.west) -- ([xshift=-.9cm]node21.east);
\draw [-Triangle] ([xshift=.8cm]node21.west) -- ([xshift=-.8cm]node21_1.east);
\draw [-Triangle] ([xshift=.8cm]node21_1.west) -- ([xshift=-.8cm]node22.east);
\draw [-Triangle] ([xshift=.8cm]node22.west) -- ([xshift=-.8cm]node22_1.east);
\draw [-Triangle] ([xshift=1.cm]node23.west) -- ([xshift=-1.cm]node23_1.east);
\draw [-Triangle] ([xshift=1.cm]node23_1.west) -- ([xshift=-.7cm]node24.east);
\draw [-Triangle] ([xshift=.8cm]node24.west) -- ([xshift=-.8cm]node24_1.east);

%\draw [loosely dashed] ([xshift=.5cm]node12.east) -- ([xshift=-.5cm]node13.west) node[] {}; 
\draw [loosely dashed] ([xshift=.1cm]node22_1.east) -- ([xshift=-.2cm]node23.west) node[] {}; 
%\draw [loosely dashed] ([xshift=.5cm]node32.east) -- ([xshift=-.5cm]node33.west) node[] {}; 
%\draw [loosely dashed] ([xshift=.5cm]node42.east) -- ([xshift=-.5cm]node43.west) node[] {}; 

\draw [-Triangle] ([xshift=-1.1cm]node20_32.east) -- ([xshift=.7cm]node31.west);
\draw [loosely dashed] ([xshift=-.1cm]node20_33.west) -- ([xshift=-.1cm]node32.east);
\draw [-Triangle] ([xshift=-1.1cm]node32.east) -- ([xshift=1.2cm]node20_32.west);
%\draw [-Triangle] ([xshift=-.5cm]node34.west) -- ([xshift=.5cm]node33.east);
%%%%%%%%%%%%%%%%%%%%%%%%%%%%%%%%%%%%%%%%%%%%%%%%%%%%%%%
\draw [-Triangle] ([xshift=-.9cm]node31.east) -- ([xshift=1.1cm]node20_31.west);
\draw [-Triangle] ([xshift=-1.2cm]node33.east) -- ([xshift=1.1cm]node20_33.west);
\draw [-Triangle] ([xshift=-1.5cm]node20_34.east) -- ([xshift=1.3cm]node33.west);
\draw [-Triangle] ([xshift=-.8cm]node34.east) -- ([xshift=1.5cm]node20_34.west);
%%%%%%%%%%%%%%%%%%%%%%%%%%%%%%%%%%%%%%%%%%%%%%%%%%%%%%%%%%
%\draw [-Triangle] ([yshift=-2mm]node11.south) -- ([yshift=2mm]node21.north);
%\draw [-Triangle] ([yshift=-2mm]node12.south) -- ([yshift=2mm]node22.north);
%\draw [-Triangle] ([yshift=-2mm]node13.south) -- ([yshift=2mm]node23.north);
%\draw [-Triangle] ([yshift=-2mm]node14.south) -- ([yshift=2mm]node24.north);

%\draw [-Triangle] ([yshift=-2mm]node31.south) -- ([yshift=2mm]node41.north);
%\draw [-Triangle] ([yshift=-2mm]node32.south) -- ([yshift=2mm]node42.north);
%\draw [-Triangle] ([yshift=-2mm]node33.south) -- ([yshift=2mm]node43.north);
%\draw [-Triangle] ([yshift=-2mm]node34.south) -- ([yshift=2mm]node44.north);

\node[below right = 1.7cm and -10.cm of m1, align=center] {Gradient Values};
%\node[above right = .7cm and -11cm of m1, align=center] {Parameter Values};

\node[below right = 0mm and -12mm of node22, align=center] (textforward) {Forward};
\node[above right = 0mm and 1mm of node32, align=center] {Backward};

\node[rounded corners, draw, below right = .8cm and 2cm of node24] (nodec) {Loss : $J(.)$};
\node[above left = 2cm and -1.6cm of nodec.west] (nodey) {$y$};
\draw [-Triangle] ([yshift=-2mm]nodey.south) -- ([yshift=2mm, xshift=5mm]nodec.north);

\path (node24_1.east)edge [-Triangle,out=0,in=90] node [midway, anchor=center] {} (nodec.north);

\path (nodec.south)edge [-Triangle,out=290,in=0] node [midway, anchor=center] {} (node34.east);

\end{tikzpicture}
\vspace*{3.5cm}
\caption[ ]{Mapping diagram associated to the entire back-propagation process.}
	\label{fig:forward-backward-pass}
\end{figure}
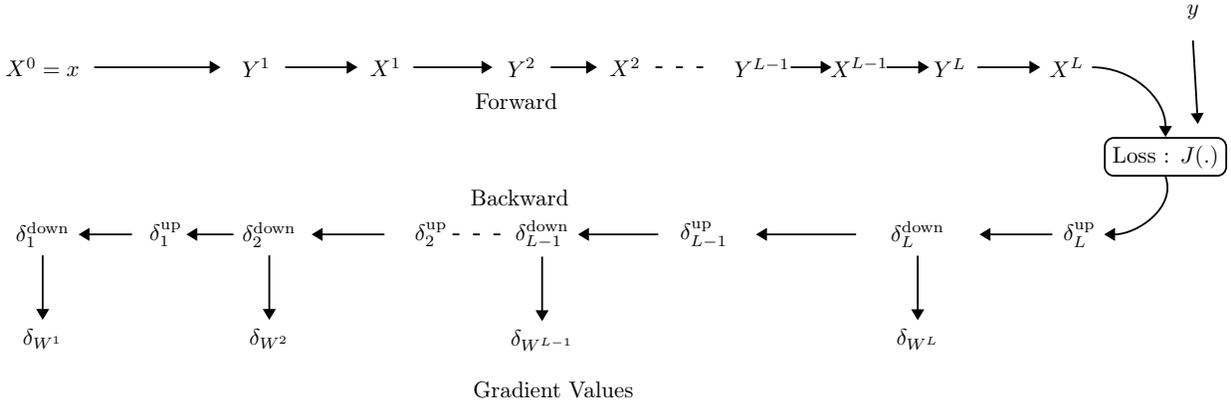
%%%%%%%%%%%%%%%%%%%%%%%%%%%%%%%%%%%%%%%%%%%%%%%%%%%%%%%%%%%%%%

\section{Main mathematical result}

~~~In this section we state our main result in the following Proposition.
\begin{proposition}[The gradient backward   propagation]
{}\noindent\label{prop}

Let $L$ be the depth of a Deep Neural Network and $N_h$ the
number of neurons in the $h$-th hidden layer. We denote by $X^0 \in \mathbb{R}^{N_0}$ the
inputs of the network, $W^h \in \mathbb{R}^{{N_{h}\times{N_{h-1}}}}$ the weights matrix
defining the synaptic strengths between the hidden layer $h$ and its
preceding $h-1$. The output $Y^h$ of the hidden layer $h$ are thus defined as follows:
%%%%%%%%%%%%%%%%%%%%%%%%%%%%%%%%%%%%%%%%%%%%%%%%%%%%%%%%%%%%%%%
 
\begin{equation}\label{fp}
X^0   =   x,\
Y^h  =   W^hX^{h-1},\
X^h   =  \sigma(Y^h),\
h   =   1,2,\cdots, L.
\end{equation}
Where $\sigma(.)$ is
a point-wise differentiable activation function. We will thus
denote by $\sigma'(.)$  its first order derivative,  $x\in \mathbb{R}^{N_0}$ is the input to the network and $W^h$ is the weight matrix
of the $h$-th layer. To optimize the neural network, we compute the partial derivatives of the loss $J(f (x), y)$ w.r.t. the weight matrices $\frac{\partial J(f (x),y)}{\partial W^h}$, with $f(x)$ and $y$ are the output of the DNN and the associated target/label respectively. This quantity can be computed similarly by  the following two-step rule:
\begin{equation}\label{bp} 
\delta^{\mathrm{up}}_{L}   =  \displaystyle\frac{\partial J(f (x),y)}{\partial X^L},\ 
\delta^{\mathrm{down}}_{h}  =   \displaystyle \delta^{\mathrm{up}}_{h}\odot \sigma'(Y^h),\  
\delta^{\mathrm{up}}_{h-1}   =  \displaystyle\left(W^h\right)^\top\delta^{\mathrm{down}}_{h},\ 
h   =    L,\cdots, 1.
 \end{equation}
Once $\delta^{\mathrm{down}}_{h}$ is computed, the weights update can be computed as

\begin{equation}\label{grad1}
\frac{\partial J(f (x),y)}{\partial W^h}=\delta^{\mathrm{down}}_{h}\left(X^{h-1} \right)^\top.
\end{equation} 
\end{proposition} 
%%%%%%%%%%%%%%%%%%%%%%%%%%%%%%%%%%%%%%%%%%%%%%%%%%%%%%%%%%%%%%%%%%%%%%%%%%%%%%%%%%%%%%%%%%%%%%%%%
\subsection*{Proof of the Proposition \ref{prop}}
~~~Firstly, for any $h=1,\cdots, L$ let us recall the   simplified notations introduced by \eqref{notations}:

 $$\delta^{\mathrm{up}}_{h}=\frac{\partial J(f (x),y)}{\partial {X^h}},\ \delta^{\mathrm{down}}_{h}=\frac{\partial J(f (x),y)}{\partial {Y^h}}.$$ Secondly, for fixed $h\in\{1,\cdots, L\}$,
  $X^h=\sigma(Y^h)$, then   \eqref{chain:2} implies that
 \begin{equation*}\label{p1}
 \frac{\partial J(f (x),y)}{\partial Y^h}=\frac{\partial J(f (x),y)}{\partial X^h}\odot\sigma'(Y^h)
\end{equation*}
thus
\begin{equation}\label{p2}
 \delta^{\mathrm{down}}_{h}=\delta^{\mathrm{up}}_{h}\odot\sigma'(Y^h).
\end{equation}
On the other hand, $Y^h=W^hX^{h-1}$, thus 
\begin{equation*}\label{p4}
\frac{\partial J(f (x),y)}{\partial X^{h-1}}=(W^h)^\top\frac{\partial J(f (x),y)}{\partial Y^h} 
\end{equation*}
by vertue of \eqref{chain:}. As consequence,

\begin{equation}\label{p5}
\delta^{\mathrm{up}}_{h-1}=(W^h)^\top\delta^{\mathrm{down}}_{h}. 
\end{equation}
Equations \eqref{p2} and \eqref{p5} implies  immediately  \eqref{bp}. Moreover, we apply again  \eqref{chain:} to the cost function $J$ and the relation 
$Y^h=W^hX^{h-1}$, we deduce that
\begin{equation*}\label{p6}
\frac{\partial J(f (x),y)}{\partial W^{h}}=\frac{\partial J(f (x),y)}{\partial Y^h}(X^{h-1})^\top=\delta^{\mathrm{down}}_{h}(X^{h-1})^\top.
\end{equation*}
 This end the proof of the Proposition \ref{prop}. Furthermore, in the practical setting mentioned in  Remark \ref{rem1}, one should  replace ${W^{h}}$ by ${W_\sharp^{h}}$ by vertue of \eqref{usefl-w-sharpe}. Thus,  we have for all 
$
h \in\{L,\cdots, 2\}$

\begin{equation*}
\label{tsr2}
\delta_{h-1}^{\mathrm{up}} =  \displaystyle({W_\sharp^{h}})^\top \delta_{h}^{\mathrm{down}},
\end{equation*}
 and then the associated two-step rule is  given by
 \begin{equation}\label{bp2} 
\delta^{\mathrm{up}}_{L}   =  \displaystyle\frac{\partial J(f (x),y)}{\partial X^L},\ 
\delta^{\mathrm{down}}_{h}  =   \displaystyle \delta^{\mathrm{up}}_{h}\odot \sigma'(Y^h),\  
\delta^{\mathrm{up}}_{h-1}   =  \displaystyle\left(W_\sharp^h\right)^\top\delta^{\mathrm{down}}_{h},\ 
h   =    L,\cdots, 1.
 \end{equation}
%%%%%%%%%%%%%%%%%%%%%%%%%%%%%%%%%%%%%%%%%%%%%%%%%%%%%%%%%%%%%%%%%%%%%%%%%%%%%%%%%%%%%%%%%%%%%%%%%%%
\section{Application to  the two simplest cases   $A[1, 1, 1]$ and $A[1, 2, 1]$}
%%%%%%%%%%%%%%%%%%%%%%%%%%%%%%%%%%
~~~The present section shows that in the following  four simplest cases associated to the DNN $A[1, N_1, 1]$ with $N_1=1,2$, we shall apply the two-step rule for back-propagation to compute the partial derivative of the elementary cost function $J$ defined by $J(f(x),y)=f(x)-y$ for any real $x$ and fixed real $y$. In this particular setting, we have the two simplest cases $A[1, 1, 1]$ and $A[1, 2, 1]$:  one neuron and two neurons in the hidden layer (see Figures \ref{fig:case1} and  \ref{fig:case3}). 
\subsection{The first case: $A[1, 1, 1]$}
~~~   The first simplest case corresponds to  
$A[1, 1, 1]$ architecture is shows by the Figure \ref{fig:case1}. Let us denote by a $W^{1}=\begin{pmatrix}
\alpha_{11}^{1} &\alpha_{12}^{1}
\end{pmatrix}$ and   $W^{2}=\begin{pmatrix}
\alpha_{1}^{2} &\alpha_{2}^{2}
\end{pmatrix}$ the weights in the first and second layer. We will evaluate the $\delta_{W^1}$ and $\delta_{W^2}$ by the differential calculus rules firstly and then recover this result by the two-step rule for back-propagation.
%%%%%%%%%%%%%%%%%%%%%%%%%%%%%%%%%%%%%%%%%%%%%%%%%%%%%%%%%%%%%%%
\tikzstyle{inputNode}=[draw,circle,minimum size=30pt,inner sep=0pt]
\tikzstyle{stateTransition}=[-stealth, thick]
\begin{figure}[htp]
	\centering

\begin{tikzpicture}[scale=1.3]
	\node[inputNode, thick] (i1) at (6, 0.5) {$X_1^0\!=\!x$};	
	\node[inputNode, thick] (i3) at (6, -0.5) {$X_2^0\!=\!1$};	
	\node[inputNode, thick] (h1) at (8, 0) {$X_1^1$};
	\node[inputNode, thick] (h6) at (8, -1.2) {$X_2^1\!=\!1$};	
	\node[inputNode, thick] (o1) at (10, 0) {$X^2$};
		%\draw[stateTransition] (5, 0.75) -- node[above] {$x$} (i1);	
	%\draw[stateTransition] (5, -0.75) -- node[above] {$1$} (i3);	
	\draw[stateTransition] (i1) -- (h1) node [midway, sloped, above] {$\alpha_{11}^{1}$};	
		\draw[stateTransition] (i3) -- (h1) node [midway, sloped, above] {$\alpha_{12}^{1}$};
	\draw[stateTransition] (h1) -- (o1) node [midway, sloped, above] {$\alpha_{1}^{2}$};
	\draw[stateTransition] (h6) -- (o1) node [midway, sloped, above] {$\alpha_{2}^{2}$};	
	
	\node[above=5pt of i1, align=center] (i1) {Input \\ layer};
	\node[above=25pt of h1, align=center] (h1) {Hidden \\ layer};
	\node[above=25pt of o1, align=center] (o1) {Output \\ layer};
	
	%\draw[stateTransition] (10.4,0) -- node[above] {$\quad X^2$} (11, 0);
	
\end{tikzpicture}

\caption[Case 1.]{The DNN associated to the case $1$.}
	\label{fig:case1}
\end{figure}
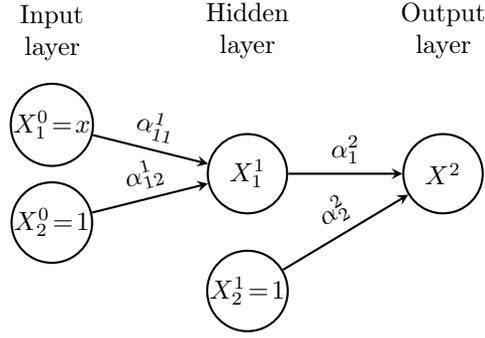
%%%%%%%%%%%%%%%%%%%%%%%%%%%%%%%%%%%%%%%%%%%%%%%%%%%%%%%%%%%%%%%%%%%%
%%%%%%%%%%%%%%%%%%%%%%%%%%%%%%%%%%%%%%%%%%%%%%%%%%%%%%%%%%%%%%%%%%%%
%%%%%%%%%%%%%%%%%%%%%%%%%%%%%%%%%%%%%%%%%%%%%%%%%%%%%%%%%%%%%%%%%%%%
\subsubsection*{The two-step forward pass :}
 \begin{align*}
                           { 
X^{0}=\begin{pmatrix}x\\
1 \end{pmatrix}
}
&
{\xmapsto[\hskip1em\sigma\hskip1em]{W^1}  
\begin{Bmatrix}
Y^1={W^1}X^{0}=\alpha_{11}^{1}x+\alpha_{12}^{1} \\
X^1=\begin{pmatrix}
\sigma(Y^1)\\
1
\end{pmatrix}=\begin{pmatrix}\sigma(\alpha_{11}^{1}x+\alpha_{12}^{1})\\
1
\end{pmatrix}
\end{Bmatrix}
} 
{\xmapsto[\hskip1em\sigma\hskip1em]{W^2}  
\begin{Bmatrix}
Y^2={W^2}X^{1}=\alpha_{1}^{2}\sigma(\alpha_{11}^{1}x+\alpha_{12}^{1})+\alpha_{2}^{2}\\
X^2=\sigma(Y^2)=\sigma\Bigl[\alpha_{1}^{2}\sigma(\alpha_{11}^{1}x+\alpha_{12}^{1})+\alpha_{2}^{2}\Bigr]
\end{Bmatrix}
}
\end{align*}
%%%%%%%%%%%%%%%%%%%%%%%%%%%%%%%%%%%%%%%%%%%%%%%%%%%%%%%%%%%%%
Hence, by using the differential calculus rules one gets

$$
 \delta_{W^2}=\begin{pmatrix}{\sigma'\Bigl[\alpha_{1}^{2}\sigma(w^1)+\alpha_{2}^{2}\Bigr]}\sigma(w^1)\\[8pt]
 {\sigma'\Bigl[\alpha_{1}^{2}\sigma(w^1)+\alpha_{2}^{2}\Bigr]}
\end{pmatrix}^\top,\  w^1:=\alpha_{11}^{1}x+\alpha_{12}^{1}
\ \mathrm{and}\ 
\delta_{W^1}=\begin{pmatrix}
{\sigma'\Bigl[\alpha_{1}^{2}\sigma(w^1)+\alpha_{2}^{2}\Bigr]}{\sigma'(w^1)}\alpha_{1}^{2}x\\[8pt]
{\sigma'\Bigl[\alpha_{1}^{2}\sigma(w^1)+\alpha_{2}^{2}\Bigr]}\sigma'(w^1)\alpha_{1}^{2}
\end{pmatrix}^\top.$$
%%%%%%%%%%%%%%%%%%%%%%%%%%%%%%%%%%%%%%%%%%%%%%%%%%%%%%%%%%%%%%%%
\subsubsection*{The two-step backward pass :}
%%%%%%%%%%%%%%%%%%%%%%%%%%%%%%%%%%%%%%%%%%%%%%%%%%%%%%%%%%%%%%%%

\begin{align*}
 \alpha_1^2{\sigma'\Bigl[\alpha_1^2\sigma(w^1)+\alpha_2^2\Bigr]}{\sigma'(w^1)}=\delta^{\mathrm{down}}_{1}
&\xmapsfrom[]{\sigma'}   
\begin{Bmatrix}
{\sigma'\Bigl[\alpha_1^2\sigma(w^1)+\alpha_2^2\Bigr]}=\delta^{\mathrm{up}}_{2}\odot{\sigma'(Y^2)}=\delta^{\mathrm{down}}_{2}\\[8pt]
\alpha_1^2{\sigma'\Bigl[\alpha_1^2\sigma(w^1)+\alpha_2^2\Bigr]}=(W_\sharp^2)^\top\delta^{\mathrm{down}}_{2}=\delta^{\mathrm{up}}_{1}
\end{Bmatrix}
\xmapsfrom[(W_\sharp^{2})^\top]{\sigma'}   
\delta^{\mathrm{up}}_{2}=1
\end{align*}
%%%%%%%%%%%%%%%%%%%%%%%%%%%%%%%%%%%%%%%%%%%%%%%%%%%%%%%%%%%%%%%
Hence, by using the  two-step rule \eqref{bp2} one gets

$$
 \delta_{W^2}=\delta^{\mathrm{down}}_{2}(X^1)^\top =\begin{pmatrix}{\sigma'\Bigl[\alpha_{1}^{2}\sigma(w^1)+\alpha_{2}^{2}\Bigr]}\sigma(w^1)\\[8pt]
 {\sigma'\Bigl[\alpha_{1}^{2}\sigma(w^1)+\alpha_{2}^{2}\Bigr]}
\end{pmatrix}^\top  \mathrm{and}\ 
\delta_{W^1}=\delta^{\mathrm{down}}_{1}(X^0)^\top =\begin{pmatrix}
{\sigma'\Bigl[\alpha_{1}^{2}\sigma(w^1)+\alpha_{2}^{2}\Bigr]}{\sigma'(w^1)}\alpha_{1}^{2}x\\[8pt]
{\sigma'\Bigl[\alpha_{1}^{2}\sigma(w^1)+\alpha_{2}^{2}\Bigr]}\sigma'(w^1)\alpha_{1}^{2}
\end{pmatrix}^\top.$$
%%%%%%%%%%%%%%%%%%%%%%%%%%%%%%%%%%
\subsection{The second case: $A[1, 2, 1]$}

~~~   The second simplest  case corresponds to  
$A[1, 2, 1]$ architecture is shows by the Figure \ref{fig:case3}.
In this case we have
$W^{1}=\begin{pmatrix}
\alpha_{11}^{1} &\alpha_{12}^{1}\\[5pt]
\alpha_{21}^{1}& \alpha_{22}^{1}
\end{pmatrix}$ and   $W^{2}=\begin{pmatrix}
\alpha_{1}^{2} &\alpha_{2}^{2}& \alpha_{3}^{2}
\end{pmatrix}$. Thus, one deduce immediately that 
%%%%%%%%%%%%%%%%%%%%%%%%%%%%%%%%%%%%%%%%%%%%%%%%%%%%%%%%%%%%%%%
 
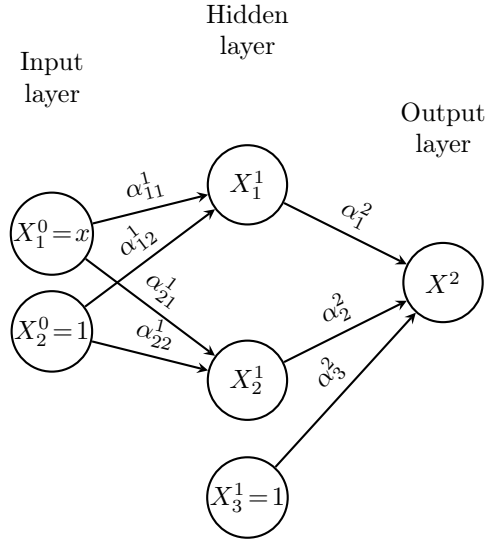
\begin{figure}[htp]
	\centering

\begin{tikzpicture}[scale=1.3]
	\node[inputNode, thick] (i1) at (6, 0.5) {$X_1^0\!=\!x$};
	
	\node[inputNode, thick] (i3) at (6, -0.5) {$X_2^0\!=\!1$};
	
	\node[inputNode, thick] (h1) at (8, 1) {$X_1^1$};
	
	\node[inputNode, thick] (h5) at (8, -1) {$X_2^1$};
	\node[inputNode, thick] (h6) at (8, -2.2) {$X_3^1\!=\!1$};
	
	\node[inputNode, thick] (o1) at (10, 0) {$X^2$};

	%\draw[stateTransition] (5, 0.75) -- node[above] {$x$} (i1);
	
	%\draw[stateTransition] (5, -0.75) -- node[above] {$1$} (i3);
	
	\draw[stateTransition] (i1) -- (h1) node [midway, sloped, above] {$\alpha_{11}^{1}$};
	
	\draw[stateTransition] (i1) -- (h5) node [midway, sloped, above] {$\alpha_{21}^{1}$};
		\draw[stateTransition] (i3) -- (h1) node [midway, sloped, above] {$\alpha_{12}^{1}$};
		\draw[stateTransition] (i3) -- (h5) node [midway, sloped, above] {$\alpha_{22}^{1}$};
	
	\draw[stateTransition] (h1) -- (o1) node [midway, sloped, above] {$\alpha_{1}^{2}$};
		\draw[stateTransition] (h5) -- (o1) node [midway, sloped, above] {$\alpha_{2}^{2}$};
	\draw[stateTransition] (h6) -- (o1) node [midway, sloped, above] {$\alpha_{3}^{2}$};

	\node[above=of i1, align=center] (l1) {Input \\ layer};
	\node[above=of h1, align=center] (h1) {Hidden \\ layer};
	\node[above=of o1, align=center] (o1) {Output \\ layer};
	
	%\draw[stateTransition] (10.4,0) -- node[above] {$\quad X^2$} (11, 0);
	
\end{tikzpicture}

\caption[Case 3.]{The  DNN associated to the case $2$.}
	\label{fig:case3}
\end{figure}
%%%%%%%%%%%%%%%%%%%%%%%%%%%%%%%%%%%%%%%%%%%%%%%%%%%%%%%%%%%%%%%%%%%%

%%%%%%%%%%%%%%%%%%%%%%%%%%%%%%%%%%%%%%%%%%%%%%%%%%%%%%%%%%%%%%%%
 %%%%%%%%%%%%%%%%%%%%%%%%%%%%%%%%%%%%%%%%%%%%%%%%%%%%%%%%%%%%%%%%%%%%
\subsubsection*{The two-step forward pass :}
 \begin{align*}
{ 
X^{0}=\begin{pmatrix}x\\
1 \end{pmatrix}
}
&
{\xmapsto[\hskip1em\sigma\hskip1em]{W^1}  
\begin{Bmatrix}
Y^1= \begin{pmatrix}\alpha_{11}^{1}x+\alpha_{12}^{1}\\ 
\alpha_{21}^{1}x+\alpha_{12}^{1}\end{pmatrix} \\[10pt]
X^1 =\begin{pmatrix}\sigma(\alpha_{11}^{1}x+\alpha_{12}^{1})\\ 
\sigma(\alpha_{21}^{1}x+\alpha_{22}^{1})\\ 
1
\end{pmatrix}
\end{Bmatrix}
} 
{\xmapsto[\hskip1em\sigma\hskip1em]{W^2}  
\begin{Bmatrix}
Y^2 =\alpha_{1}^{2}\sigma(\alpha_{11}^{1}x+\alpha_{12}^{1})+\alpha_{2}^{2}\sigma(\alpha_{21}^{1}x+\alpha_{22}^{1})+\alpha_{3}^{2}
\\[7pt]
X^2 =
\sigma\Bigl[\alpha_{1}^{2}\sigma(\alpha_{11}^{1}x+\alpha_{12}^{1})+\alpha_{2}^{2}\sigma(\alpha_{21}^{1}x+\alpha_{22}^{1})+\alpha_{3}^{2}\Bigr]
\end{Bmatrix}
}
\end{align*}
Hence, by using the differential calculus rules one gets

$$
 \delta_{W^2}=\begin{pmatrix} {\sigma'\Bigl[\alpha_{1}^{2}\sigma(w^1)+\alpha_{2}^{2}\sigma(w^2)+\alpha_{3}^{2}\Bigr]}\sigma(w^1)\\[7pt]
 {\sigma'\Bigl[\alpha_{1}^{2}\sigma(w^1)+\alpha_{2}^{2}\sigma(w^2)+\alpha_{3}^{2}\Bigr]}\sigma(w^2)\\[7pt]
 {\sigma'\Bigl[\alpha_{1}^{2}\sigma(w^1)+\alpha_{2}^{2}\sigma(w^2)+\alpha_{3}^{2}\Bigr]}
\end{pmatrix}^\top,\ w^1:=\alpha_{11}^{1}x+\alpha_{12}^{1}, w^2:=\alpha_{21}^{1}x+\alpha_{22}^{1}$$
and
$$
\delta_{W^1}=\begin{pmatrix}\alpha_1^2x{\sigma'(w^1)}{\sigma'\Bigl[\alpha_{1}^{2}\sigma(w^1)+\alpha_{2}^{2}\sigma(w^2)+\alpha_{3}^{2}\Bigr]} & \alpha_1^2{\sigma'(w^1)}{\sigma'\Bigl[\alpha_{1}^{2}\sigma(w^1)+\alpha_{2}^{2}\sigma(w^2)+\alpha_{3}^{2}\Bigr]}\\[7pt]
\alpha_2^2x{\sigma'(w^2)}{\sigma'\Bigl[\alpha_{1}^{2}\sigma(w^1)+\alpha_{2}^{2}\sigma(w^2)+\alpha_{3}^{2}\Bigr]} & \alpha_2^2{\sigma'(w^2)}{\sigma'\Bigl[\alpha_{1}^{2}\sigma(w^1)+\alpha_{2}^{2}\sigma(w^2)+\alpha_{3}^{2}\Bigr]}
\end{pmatrix}.$$
%%%%%%%%%%%%%%%%%%%%%%%%%%%%%%%%%%%%%%%%%%%%%%%%%%%%%%%%%%%%%
\subsubsection*{The two-step backward pass :}
%%%%%%%%%%%%%%%%%%%%%%%%%%%%%%%%%%%%%%%%%%%%%%%%%%%%%%%%%%%%%%%%

\begin{align*}
%
%\begin{Bmatrix}\!\!
\!\begin{pmatrix} \alpha_1^2\sigma'\Bigl[\alpha_{1}^{2}\sigma(w^1)+\alpha_{2}^{2}\sigma(w^2)+\alpha_{3}^{2}\Bigr]\sigma'(w^1)\\[8pt]
\alpha_2^2\sigma'\Bigl[\alpha_{1}^{2}\sigma(w^1)+\alpha_{2}^{2}\sigma(w^2)+\alpha_{3}^{2}\Bigr]\sigma'(w^2)
\end{pmatrix}\!\!=\!\delta^{\mathrm{down}}_{1}
%\end{Bmatrix}\!\!\! 
&\!\xmapsfrom[]{\sigma'}\!\! 
\begin{Bmatrix}\!\!\!\!
{\sigma'\Bigl[\alpha_{1}^{2}\sigma(w^1x)+\alpha_{2}^{2}\sigma(w^2)+\alpha_{3}^{2}\Bigr]}\!=\!\delta^{\mathrm{down}}_{2}\\[8pt]
\!\!\!\begin{pmatrix} \alpha_1^2\sigma'\Bigl[\alpha_{1}^{2}\sigma(w^1)+\alpha_{2}^{2}\sigma(w^2)+\alpha_{3}^{2}\Bigr]\\[8pt]
\alpha_2^2\sigma'\Bigl[\alpha_{1}^{2}\sigma(w^1)+\alpha_{2}^{2}\sigma(w^2)+\alpha_{3}^{2}\Bigr]
\!\end{pmatrix}\!\!=\!\delta^{\mathrm{up}}_{1}
\!\end{Bmatrix}\!\!
 \xmapsfrom[(W_\sharp^{2})^\top]{\sigma'}  \!  
\delta^{\mathrm{up}}_{2}\!=\!\!1
\end{align*}
Hence, by using the  two-step rule \eqref{bp2} one gets

$$
 \delta_{W^2}=\delta_2^{\mathrm{down}}(X^1)^\top=\begin{pmatrix} {\sigma'\Bigl[\alpha_{1}^{2}\sigma(w^1)+\alpha_{2}^{2}\sigma(w^2)+\alpha_{3}^{2}\Bigr]}\sigma(w^1)\\[8pt]
 {\sigma'\Bigl[\alpha_{1}^{2}\sigma(w^1)+\alpha_{2}^{2}\sigma(w^2)+\alpha_{3}^{2}\Bigr]}\sigma(w^2)\\[8pt]
 {\sigma'\Bigl[\alpha_{1}^{2}\sigma(w^1)+\alpha_{2}^{2}\sigma(w^2)+\alpha_{3}^{2}\Bigr]}
\end{pmatrix}^\top$$
and
$$
\delta_{W^1}=\delta_1^{\mathrm{down}}(X^0)^\top=\begin{pmatrix}\alpha_1^2x{\sigma'(w^1)}{\sigma'\Bigl[\alpha_{1}^{2}\sigma(w^1)+\alpha_{2}^{2}\sigma(w^2)+\alpha_{3}^{2}\Bigr]} & \alpha_1^2{\sigma'(w^1)}{\sigma'\Bigl[\alpha_{1}^{2}\sigma(w^1)+\alpha_{2}^{2}\sigma(w^2)+\alpha_{3}^{2}\Bigr]}\\[8pt]
\alpha_2^2x{\sigma'(w^2)}{\sigma'\Bigl[\alpha_{1}^{2}\sigma(w^1)+\alpha_{2}^{2}\sigma(w^2)+\alpha_{3}^{2}\Bigr]} & \alpha_2^2{\sigma'(w^2)}{\sigma'\Bigl[\alpha_{1}^{2}\sigma(w^1)+\alpha_{2}^{2}\sigma(w^2)+\alpha_{3}^{2}\Bigr]}
\end{pmatrix}.$$
%%%%%%%%%%%%%%%%%%%%%%%%%%%%%%%%%%
%%%%%%%%%%%%%%%%%%%%%%%%%%%%%%%%%%

%%%%%%%%%%%%%%%%%%%%%%%%%%%%%%%%%%%%%%%%%%%%%%%%%%%%%%%%%%%%%%%%%%%%%%% 
%%%%%%%%%%%%%%%%%%%%%%%%%%%%%%%%%%%%%%%%%%%%%%%%%%%%%%%%%%%%%%%%%%%%
%%%%%%%%%%%%%%%%%%%%%%%%%%%%%%%%%%
\section{Related works and Conclusion}
~~~ It is interesting to cite related works which have some connections
with ours.  To the best of our knowledge, in literature, the related  works to this paper  are \cite{Alber2018} and \cite{Hojabr2020}. In particular, in the first paper the authors uses some decomposition of the partial derivatives of the cost function, similar to the two-step formula \eqref{delta-X}, to replace the  standard  back-propagation. In addition, they show (experimentally) that for specific scenarios, the two-step decomposition yield better generalization performance than the   one based on the  standard  back-propagation. But in the second article, the authors find some similar update equation similar to the one given by \eqref{delta-X} that report similarly to standard back-propagation at convergence. Moreover, this method discovers new variations of the back-propagation
 by learning new propagation rules that optimize the generalization performance after a few epochs of training.

In conclusion, we have provided a two-step rule for back-propagation similar to the one for forward propagation. We hope that it  serve as a pedagogical material for data scientists, and may also inspire the exploration of novel   approaches for optimizing some artificial neural networks training algorithms. As future work, we envision to explore some experimental  issues to compare the performance of this two-step rule for back-propagation and the standard one.

\subsection*{Declarations}
\noindent\textbf{Conflicts of interest/Competing interests} No  conflict of interest.

\end{document}